\documentclass[10pt,twocolumn,letterpaper]{article}
\usepackage[T1]{fontenc}
\usepackage[hmargin=0.8in,vmargin=0.85in,columnsep=0.3in]{geometry}
\usepackage{times}
\usepackage{graphicx}
\usepackage{xcolor}
\usepackage{amssymb}
\usepackage{booktabs}
\usepackage[numbers,sort&compress]{natbib}
\usepackage[font=small,labelsep=period]{caption}
\usepackage{etoolbox}
\ifdefined\pdfinclusioncopyfonts
\fi

\usepackage{algorithm}
\usepackage{algpseudocode}
\usepackage{mathtools}
\usepackage{tabularx}
\usepackage{multirow}
\usepackage{placeins}

\newcommand{\kvbestrow}[1]{\kvsetbestbits#1\relax}
\def\kvsetbestbits#1#2#3#4#5#6#7#8#9\relax{
  \expandafter\gdef\csname kvbest1\endcsname{#1}
  \expandafter\gdef\csname kvbest2\endcsname{#2}
  \expandafter\gdef\csname kvbest3\endcsname{#3}
  \expandafter\gdef\csname kvbest4\endcsname{#4}
  \expandafter\gdef\csname kvbest5\endcsname{#5}
  \expandafter\gdef\csname kvbest6\endcsname{#6}
  \expandafter\gdef\csname kvbest7\endcsname{#7}
  \expandafter\gdef\csname kvbest8\endcsname{#8}
  \expandafter\gdef\csname kvbest9\endcsname{#9}
}
\newcommand{\kvbestfont}[1]{\ifnum\csname kvbest#1\endcsname=1\relax\bfseries\fi}
\kvbestrow{000000000}
\newcolumntype{K}[1]{>{\kvbestfont{#1}\centering\arraybackslash}X}
\usepackage{colortbl}
\usepackage{newfloat}
\usepackage{listings}
\definecolor{kvsoftblue}{HTML}{CFE3F6}
\definecolor{kvsoftcyan}{HTML}{CDEFF1}
\definecolor{kvsoftgray}{HTML}{F2F2F2}

\DeclareCaptionStyle{ruled}{labelfont=normalfont,labelsep=colon,strut=off}
\floatstyle{ruled}
\newfloat{listing}{tb}{lst}{}
\floatname{listing}{Listing}

\definecolor{linkblue}{rgb}{0.14,0.33,0.57}
\usepackage[breaklinks,colorlinks,allcolors=linkblue]{hyperref}

\title{CommitKV: Commit-Aligned KV Cache Compression for Multimodal Agents}
\author{
    Weizhong Huang\textsuperscript{1,\textdagger}\quad
    Jinchao Zhang\textsuperscript{2,*}\quad
    Xiawu Zheng\textsuperscript{1,*}\\[0.5em]
    \textsuperscript{1}Xiamen University\quad
    \textsuperscript{2}WeChat AI, Tencent Inc., China\\
    \textsuperscript{*}Corresponding authors
}
\date{}
\newcommand{\method}{\textsc{CommitKV}}

\AtBeginEnvironment{thebibliography}{\small}

\begin{document}

\maketitle
\begingroup
\renewcommand{\thefootnote}{\fnsymbol{footnote}}
\footnotetext[2]{This work was done when Weizhong Huang was an intern at WeChat AI.}
\endgroup

\setcounter{dbltopnumber}{1}

\begin{abstract}
Multimodal agents gather visual evidence through reasoning and tool use, extending their ability to solve tasks beyond the initial input. However, accumulating observations increases KV cache memory and inference costs. Token eviction reduces these costs by retaining useful cached tokens. Existing methods use attention, redundancy, or stage-specific queries to rank tokens, without explicitly comparing the same history across tool interactions. When observations shift attention, they may mistake temporarily reduced influence for lost usefulness and evict information needed later. We propose \method{}, a training-free KV cache compression method that accounts for this changing evidence use. Commit-aligned measurement compares the deletion effects of the same page before a tool call completes and after its observation returns. Lifecycle-aware compression retires pages with low influence in both windows and temporarily protects pages influential before the call. Joint scoring combines attention importance, key redundancy, and token-level deletion effects to fill the remaining budget. Extensive experiments cover four open-weight multimodal models and eight benchmarks. \method{} achieves the highest average accuracy among the compared compression methods on all four models. On Qwen3-VL-8B, it achieves \textbf{69.85\%} at a 512-token budget, only \textbf{0.91\%} below FullKV, and exceeds the strongest compression baseline by \textbf{8.44\%} at 256 tokens. Deploying the Qwen3-VL-8B agent with SGLang and vLLM at a 256-token budget reduces peak allocated GPU memory by \textbf{54.37\%--55.23\%} and accelerates end-to-end inference by $\mathbf{2.12\text{--}2.17\times}$ over FullKV on the matched workload.
\end{abstract}

\begin{figure}[t]
\centering
\includegraphics[width=\columnwidth]{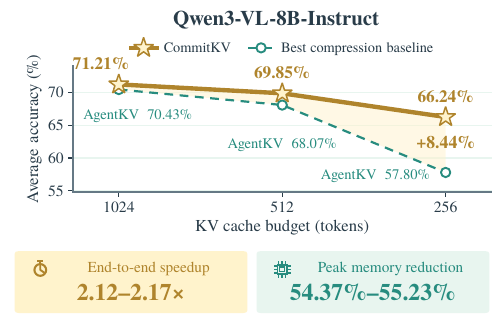}
\caption{\textbf{Qwen3-VL-8B accuracy and efficiency.} Curves show eight-benchmark means and the best compression baseline. Cards summarize vLLM/SGLang gains over each backend's FullKV at $B=256$, with 8,192 input and 163,840 generated tokens.}
\label{fig:teaser}
\end{figure}

\section{Introduction}
\label{sec:intro}

Multimodal agents use reasoning and tools for visual search, information retrieval, and complex tasks, extending their capabilities beyond the initial input~\citep{vstar2023,react}. During successive interactions, they issue tool calls based on available information and continue reasoning over returned text or image observations. Reusing key-value (KV) caches across turns avoids recomputing historical context, but accumulating observations and generated text continually expand the cache, increasing GPU memory use and inference costs~\citep{gao2024cost,sglang}. Compressing this cache while preserving information needed later is therefore important for efficient multimodal agent deployment.

Among compression methods, token eviction identifies and removes less important tokens, directly shortening the cache and reducing decoding-time memory accesses. Prior work shows that it can substantially compress caches while largely preserving long-context understanding and reasoning ability~\citep{rkv,snapkv}. Its central task is to estimate cached-token importance and retain information useful for subsequent generation within a limited budget.

\begin{figure*}[t]
\centering
\captionsetup{skip=2pt}
\includegraphics[width=\textwidth]{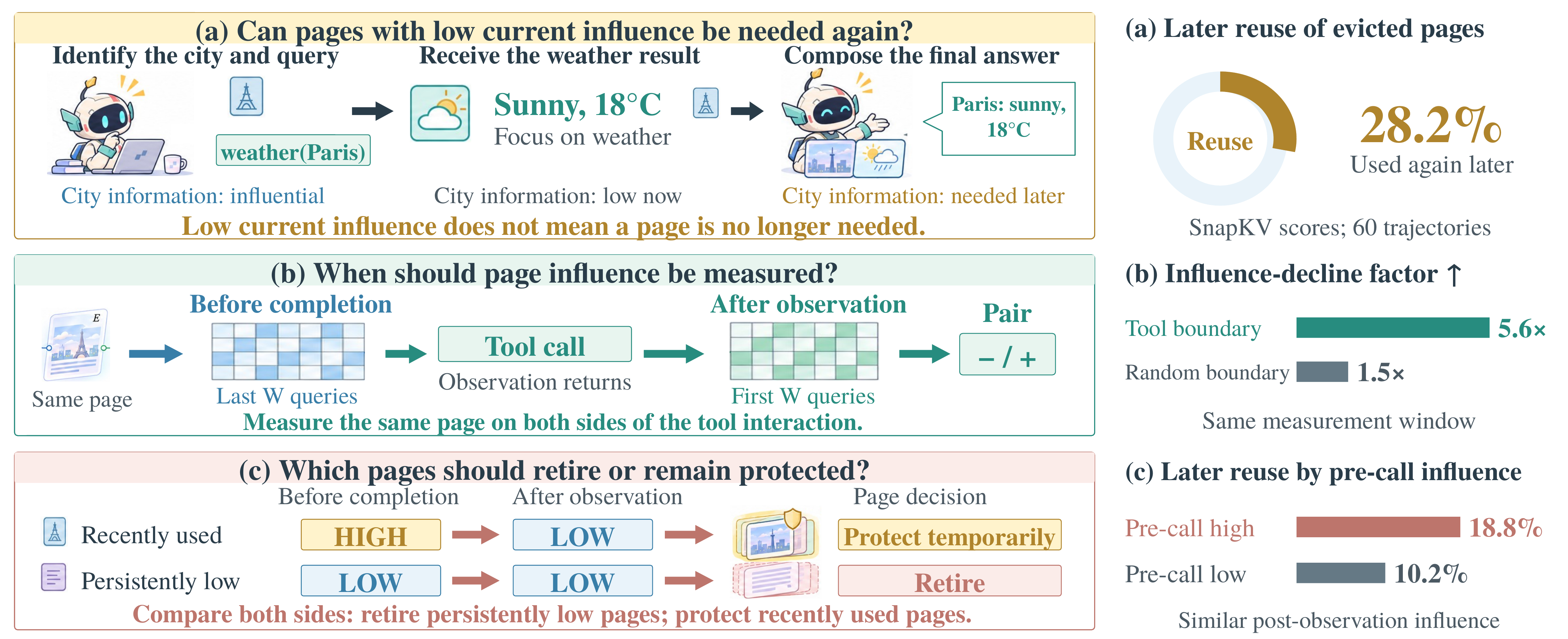}
\caption{\textbf{Why is current influence insufficient for page decisions?} (a) Paris weather example and later reuse under SnapKV scoring in FullKV replay. (b) Tool versus random boundaries motivate paired measurements. (c) Pages with similar post-observation influence have different reuse rates across pre-call groups, motivating joint retirement/protection decisions. Full analysis: Section~\ref{sec:observations}.}
\label{fig:lifecycle_motivation}
\end{figure*}

Existing methods assess token importance through attention, redundancy, or interaction phases. SnapKV~\citep{snapkv} aggregates attention over recent observation windows; R-KV~\citep{rkv} combines attention importance with key redundancy; TriAttention~\citep{triattention} uses pre-RoPE query--key geometry; and AgentKV~\citep{agentkv2026} selects representative queries from reasoning, actions, and observations. However, these methods do not explicitly compare the same historical information's role before call completion and after observation return. In the Paris weather example (Figure~\ref{fig:lifecycle_motivation}a), city information from an earlier observation determines the arguments of \texttt{weather(Paris)}. Once the weather returns, attention shifts to the weather information and the city information's current influence may fall. These methods may then mistakenly evict the city information, although the final answer still needs both the city and its weather. Current importance scores alone may therefore remove information before its next use.

We propose \method{}, a training-free KV cache compression method for multimodal agents. Commit-aligned measurement divides cached events into contiguous pages and compares each page's deletion effects before call completion and after observation return, recording its changing role across the interaction. Lifecycle-aware compression then retires pages with low influence on both sides and temporarily protects recently used pages, preventing their removal solely because a new observation shifts attention. Joint scoring subsequently combines attention importance, key redundancy, and singleton deletion effects to select the remaining tokens within budget. The procedure reuses existing attention quantities without additional training or forward passes.

We evaluate four open-weight multimodal models on eight benchmarks, covering dense, mixture-of-experts, and hybrid-attention architectures with 8B--35B parameters. The benchmarks assess tool-assisted information acquisition and use, visual perception, and reasoning. \method{} achieves the highest average accuracy among the compared compression methods on all four models. On Qwen3-VL-8B, it loses only \textbf{0.91\%} relative to FullKV at 512 tokens and exceeds the strongest compression baseline by \textbf{8.44\%} at the tighter 256-token budget. Deploying the Qwen3-VL-8B agent with vLLM and SGLang reduces peak allocated GPU memory by \textbf{54.37\%--55.23\%} and yields \textbf{2.12--2.17}$\boldsymbol{\times}$ end-to-end speedups over each backend's FullKV reference. These results demonstrate that \method{} preserves multimodal task accuracy while reducing memory use and accelerating inference.

\section{Related Work}
\label{sec:related}

\paragraph{Multimodal agents.}
Agents combine reasoning, tool calls, and environment feedback to solve multi-turn tasks~\citep{masterman2024landscape,react,zhang2025tool}. V*~\citep{vstar2023} introduces guided visual search; DeepEyes~\citep{deepeyes2025}, Mini-o3~\citep{minio32025}, and Pixel Reasoner~\citep{pixelreasoner2025} further support visual inspection and image manipulation. Search and browsing acquire external information through tools~\citep{mmsearch2024,mmbrowsecomp2025,mmsearchplus2025}, while visual environment interaction relies on successive screenshots and actions~\citep{molmoweb,osworld}. Each returned text or image becomes part of subsequent reasoning, so longer interaction chains enlarge the history accessed by later steps. These capabilities let agents acquire information as tasks unfold, but accumulating observations also continually increase KV cache costs.

\paragraph{KV cache compression.}
KV cache compression preserves useful context in a smaller cache to reduce memory use and inference costs~\citep{shi2024keep,gao2025rethinking}. Existing work improves retention through token importance, query relevance, and cache-budget allocation~\citep{pyramidkv,adakv,scissorhands,quest,streamingllm,h2o}. SnapKV~\citep{snapkv} aggregates recent attention, R-KV~\citep{rkv} combines attention importance with key redundancy, TriAttention~\citep{triattention} uses pre-RoPE query--key geometry, and AgentKV~\citep{agentkv2026} selects representative queries by interaction phase. Work on multi-turn interaction also considers cross-turn intent, semantic regions, episodic memory, and context management~\citep{intentkv,memdecay,epicache,selfgc2026}. \method{} focuses on changing historical-information roles across tool interactions, using paired measurement and lifecycle management to preserve information needed by subsequent reasoning and tool calls.

\section{Preliminaries and Motivation}
\label{sec:observation}

\subsection{Preliminaries}
\label{sec:Problem_Formulation}

\paragraph{Multimodal ReAct agents with KV reuse.}
An agent answers a query $\boldsymbol q$, which contains text and images, through repeated reasoning and tool use~\citep{react}. At turn $t$, given the history
\begin{equation}
\boldsymbol x_t=(\boldsymbol q,\boldsymbol z_1,\boldsymbol a_1,
\boldsymbol o_1,\ldots,\boldsymbol z_{t-1},\boldsymbol a_{t-1},
\boldsymbol o_{t-1}),
\label{eq:react_trace}
\end{equation}
the agent generates reasoning $\boldsymbol z_t$ and a decision $\boldsymbol a_t$, which is either a tool call or the final answer. A tool call returns an observation $\boldsymbol o_t$, which may contain text, images, or both. With cross-turn KV reuse~\citep{gao2024cost,sglang}, the server prefills only the new observation at each turn and appends the KV states of the generated $\boldsymbol z_t$ and $\boldsymbol a_t$, so the cache grows with the full trajectory.

\paragraph{KV cache compression.}
Let $\boldsymbol K,\boldsymbol V\in\mathbb R^{M\times H\times S\times d}$ denote the cached keys and values of $S$ resident tokens over $M$ layers and $H$ KV heads, where $d$ is the dimension of the key and value vectors in each KV head. Let $B$ denote the maximum number of tokens retained after each compression. At the $j$-th \emph{compression checkpoint}, a policy selects a retained index set $\mathcal I_j$ with $|\mathcal I_j|\leq B$ and keeps the corresponding keys, values, and original position indices:
\begin{equation}
\widetilde{\boldsymbol K}_j=\boldsymbol K_j[:,:,\mathcal I_j,:],\quad
\widetilde{\boldsymbol V}_j=\boldsymbol V_j[:,:,\mathcal I_j,:],\quad
|\mathcal I_j|\leq B.
\label{eq:kv_compress}
\end{equation}
Position indices are kept unchanged because RoPE~\citep{rope} and its multimodal variant M-RoPE~\citep{qwen2vl2024} encode relative positions from them.

\paragraph{Event pages.}
Grouping cached tokens into chunks or memory blocks is used in structured cache compression~\citep{chunkkv,pagedeviction}. We treat the user query $\boldsymbol q$, every completed tool call $\boldsymbol a_i$, and every returned observation $\boldsymbol o_i$ as \emph{agent events} and divide them into \emph{event pages} of at most $G$ contiguous tokens, the units at which influence is measured and lifecycle decisions are applied. Visual tokens within an observation are ordered row by row, from left to right and top to bottom, and grouped into pages of at most $G$ consecutive tokens.

\paragraph{Deletion effect.}
We measure a page's importance by the relative attention-output change caused by its removal~\citep{caote2025}. For each query $p$ and head $h$, let $\boldsymbol o$ and $\widetilde{\boldsymbol o}$ denote the outputs before and after removing page $\mathcal E$. Its deletion effect over query window $\mathcal Q$ is
\begin{equation}
\mathcal R(\mathcal E;\mathcal Q)
=\max_{p\in\mathcal Q,\,h}
\frac{\|\boldsymbol o-\widetilde{\boldsymbol o}\|_2}
     {\|\boldsymbol o\|_2}.
\label{eq:window_residual}
\end{equation}
We compare the relative output changes across all positions and attention heads in the window and take the largest value. A larger value means that removing the page has a greater effect on the output. We compute this quantity from existing attention quantities without an extra forward pass.

\subsection{Motivation}
\label{sec:motivation}
Successive tool observations accumulate history and increase agents' KV cache costs, while changing which historical information later reasoning and tool calls need. \textbf{How can we compress the cache while preserving history needed later?} We examine changing information needs, measurement timing, and page retirement and protection (Figure~\ref{fig:lifecycle_motivation}).

\paragraph{Can pages with low current influence be needed again?}
Tool interactions change information needs. In the Paris weather example (Figure~\ref{fig:lifecycle_motivation}a), earlier city information determines the arguments of \texttt{weather(Paris)}. The returned weather shifts attention away from the city page, although the final answer may still need both. Retaining the city information lets the final response connect the returned weather to the location requested by the user. In a separate analysis replaying 60 FullKV trajectories, $28.2\%$ of pages evicted according to SnapKV scores are reused later. Thus, \textbf{low current influence does not imply that a page is no longer needed; cache decisions should account for changing roles across interactions.}

\begin{figure*}[t]
\centering
\captionsetup{skip=2pt}
\includegraphics[width=\textwidth,keepaspectratio]{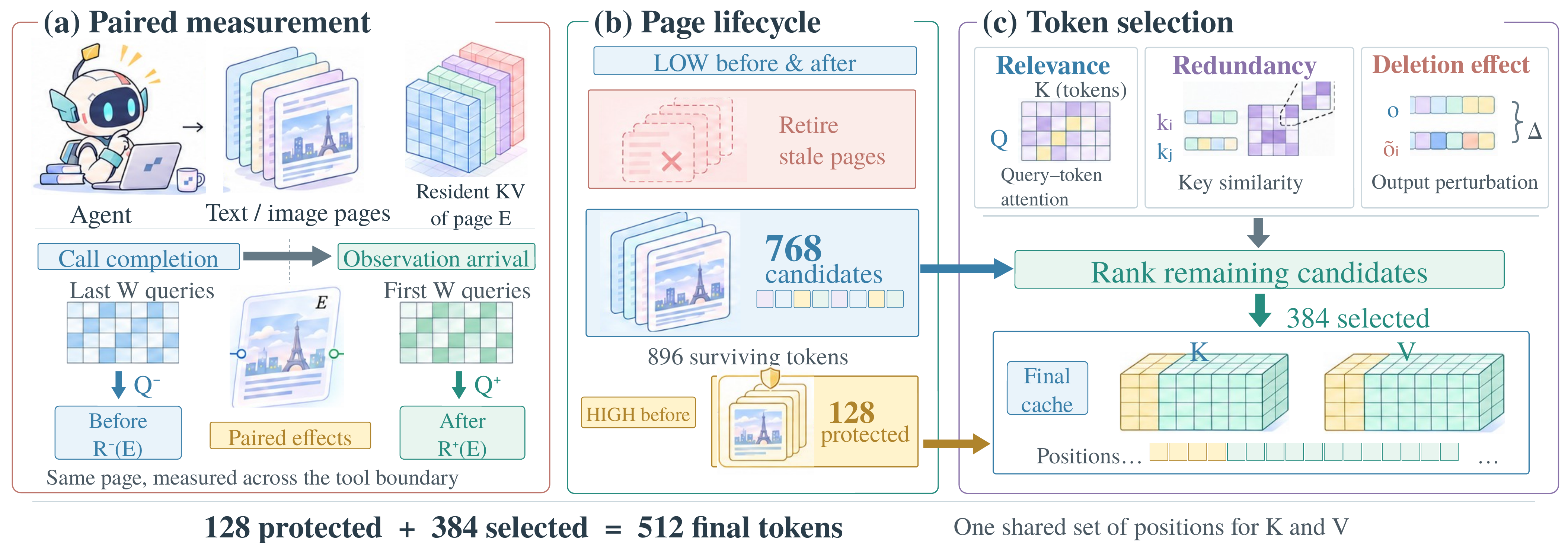}
\caption{\textbf{From event pages to a budgeted KV cache.} Paired effects guide retirement and protection; remaining candidates are ranked for the unused slots. Gold denotes protected tokens, which bypass ranking. Of 896 surviving tokens, 128 are protected and 384 of 768 candidates are selected, meeting $B=512$. Keys, values, and positions share retained indices.}
\label{fig:overview}
\end{figure*}

\paragraph{When should page influence be measured?}
Measuring only after an observation may miss a page's earlier role. Figure~\ref{fig:lifecycle_motivation}b compares influence changes around tool boundaries and random within-turn boundaries: mean influence drops by $5.6\times$ at tool boundaries, versus $1.5\times$ at random boundaries. This difference motivates recording the change around tool interactions. We therefore \textbf{measure the same page before call completion and after the observation returns, recording its role on both sides.}

\paragraph{Which pages should retire or remain protected?}
Low post-observation influence alone does not distinguish recently used pages from persistently low-influence pages. We select pages with similar post-observation influence, divide them by high or low pre-call influence, and compare later reuse (Figure~\ref{fig:lifecycle_motivation}c). Reuse rates are $18.8\%$ and $10.2\%$, respectively. Both groups look similarly unimportant after the observation, yet their earlier roles differ. This difference shows that \textbf{post-observation influence alone does not fully indicate later reuse; pre-call influence provides additional information.} We therefore combine measurements from both sides of the interaction to decide retirement and temporary protection.

These analyses motivate paired measurement in Section~\ref{sec:commit} and lifecycle management in Section~\ref{sec:transition}; Section~\ref{sec:observations} provides the complete experimental analysis.

\section{Method}
\label{sec:method}

Tool interactions change historical information needs (Section~\ref{sec:motivation}). \method{} compares event-page deletion effects before call completion and after observation return (Section~\ref{sec:commit}) to guide retirement and temporary protection (Section~\ref{sec:transition}). These page-level decisions determine eligibility and protection. Joint scoring then combines relevance and singleton deletion effects to fill the remaining token slots within $B$ (Section~\ref{sec:visual_refinement}). Figure~\ref{fig:overview} and Algorithm~\ref{alg:commitkv} summarize the procedure.

\subsection{Commit-Aligned Measurement}
\label{sec:commit}

Following Section~\ref{sec:motivation}, we measure both sides of a tool interaction. The \emph{tool-call commit} $c$ ends the generated call $\boldsymbol a_t$ at position $|\boldsymbol y_t|$, where $\boldsymbol y_t=(\boldsymbol x_t,\boldsymbol z_t,\boldsymbol a_t)$. After the observation $\boldsymbol o_t$ is prefilled, decoding of the next turn starts at position $|\boldsymbol x_{t+1}|+1$. With a window size $W$, we define query windows before call completion and after the observation returns:
\begin{equation}
\begin{aligned}
\mathcal Q_c^{-}&=\{|\boldsymbol y_t|-W+1,\ldots,|\boldsymbol y_t|\},\\
\mathcal Q_c^{+}&=\{|\boldsymbol x_{t+1}|+1,\ldots,|\boldsymbol x_{t+1}|+W\}.
\end{aligned}
\label{eq:query_windows}
\end{equation}
$\mathcal Q_c^{-}$ contains the last $W$ queries of the tool call, which reflect what the agent relied on to issue the request. $\mathcal Q_c^{+}$ contains the first $W$ queries after the observation, which reflect what the agent needs once the observation returns. For every event page $\mathcal E$ that is fully resident at the commit, \method{} records
\begin{equation}
\mathcal R_c^{-}(\mathcal E)=\mathcal R(\mathcal E;\mathcal Q_c^{-}),
\qquad
\mathcal R_c^{+}(\mathcal E)=\mathcal R(\mathcal E;\mathcal Q_c^{+}).
\label{eq:pre_post}
\end{equation}
Both windows are short, as in SnapKV~\citep{snapkv}, but lie on opposite sides of the interaction to record the same page's changing role. Comparing the same page keeps the unit of analysis fixed while exposing its change in influence.

\subsection{Lifecycle-Aware Compression}
\label{sec:transition}
\label{sec:retirement}

\begin{figure}[t]
\centering
\captionsetup{skip=2pt}
\includegraphics[width=\columnwidth]{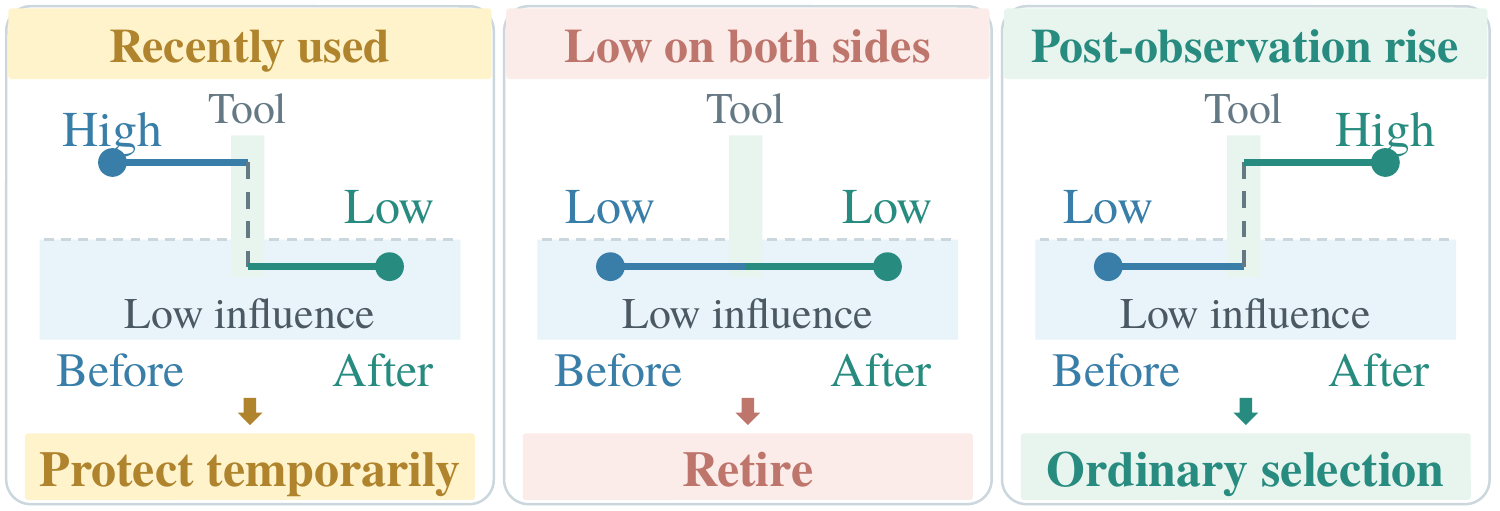}
\caption{\textbf{Why measure both sides of an interaction?} High pre-call influence marks recent use, even when influence later falls. Pages low on both sides retire; other pages compete in selection.}
\label{fig:paired_reuse}
\end{figure}

Following Section~\ref{sec:motivation}, paired effects guide retirement and temporary protection. Only pages measured in both windows ($\mathcal P_c$) are eligible for retirement. Figure~\ref{fig:paired_reuse} illustrates why a low current score alone is insufficient.

\paragraph{Retiring persistently low-influence pages.}
Persistently low-influence pages have small deletion effects in both windows. \method{} retires
\begin{equation}
\mathcal T_c=\bigl\{
\mathcal E\in\mathcal P_c \;\big|\;
\mathcal R_c^{-}(\mathcal E)\leq\tau_{\mathrm{dead}},\;
\mathcal R_c^{+}(\mathcal E)\leq\tau_{\mathrm{dead}}
\bigr\},
\label{eq:transition_rule}
\end{equation}
where $\tau_{\mathrm{dead}}$ is a deletion-effect threshold. The set $\mathcal T_c$ contains pages retired at commit $c$. We record retired pages across commits in $\mathcal D$, updating $\mathcal D\leftarrow\mathcal D\cup\mathcal T_c$. Their tokens are removed at the next compression checkpoint. Following Section~\ref{sec:motivation}, retirement considers both measurements rather than current low influence alone.

\paragraph{Protecting recently used pages.}
To temporarily retain historical information used before the call, \method{} marks
\begin{equation}
\mathcal U_c=\bigl\{\mathcal E \;\big|\; \mathcal R_c^{-}(\mathcal E)\geq\tau_{\mathrm{use}}\bigr\},
\qquad \tau_{\mathrm{use}}>\tau_{\mathrm{dead}},
\label{eq:protect_rule}
\end{equation}
as \emph{recently used}. Protection starts before call completion and spans the checkpoint after observation return. It lasts until the next measured commit $c'$, when $\mathcal U_{c'}$ replaces $\mathcal U_c$: continued high influence renews protection; otherwise it expires. Pages without a lifecycle label, including those between the thresholds, compete in ordinary token selection. When protection expires, the page's tokens can still survive if they rank highly enough in this selection.

\begin{figure}[!htb]
\centering
\captionsetup{skip=2pt}
\includegraphics[width=\columnwidth]{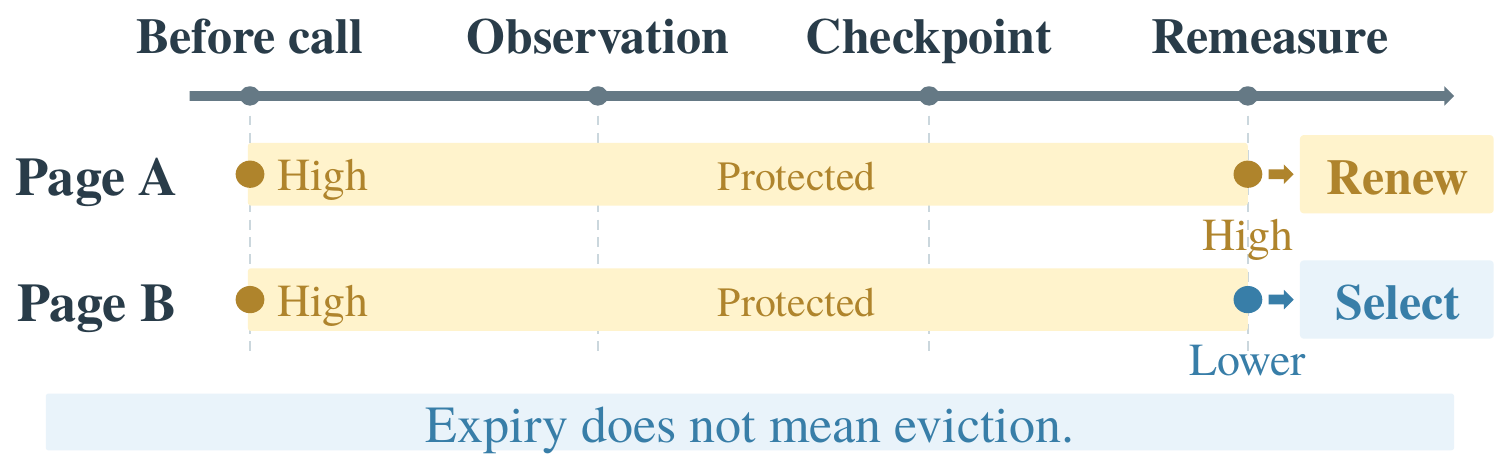}
\caption{\textbf{Temporary protection across an interaction.} Protection begins before the call and spans observation return and compression. At the next measured commit, continued high influence renews protection; otherwise the page returns to ordinary selection. Losing protection does not itself evict a page.}
\label{fig:lifecycle_illustration}
\end{figure}

Figure~\ref{fig:lifecycle_illustration} illustrates when protection is renewed or expires.

\paragraph{Cache selection.}
At checkpoint $j$, we remove tokens in retired pages $\mathcal D$ (Eq.~\eqref{eq:transition_rule}). Among pages with active protection (Eq.~\eqref{eq:protect_rule}), larger pre-call effects receive priority; $\mathcal N_j$ denotes the indices actually protected. Section~\ref{sec:visual_refinement} selects other tokens to form $\mathcal I_j$ (Section~\ref{sec:Problem_Formulation}). Both subsets share budget $B$:
\begin{equation}
 \mathcal N_j\subseteq\mathcal I_j,
 \qquad |\mathcal I_j|\leq B.
\label{eq:retained_indices}
\end{equation}

\subsection{Joint Scoring and Token Selection}
\label{sec:visual_refinement}

Section~\ref{sec:transition} removes retired pages and determines the protected token set $\mathcal N_j$. If the surviving cache exceeds budget $B$, we retain $\mathcal N_j$ and rank the other candidates to fill the remaining $B-|\mathcal N_j|$ slots. The final score combines attention importance, key redundancy, and deletion effects.

\begin{figure}[t]
\centering
\captionsetup{skip=2pt}
\includegraphics[width=\columnwidth]{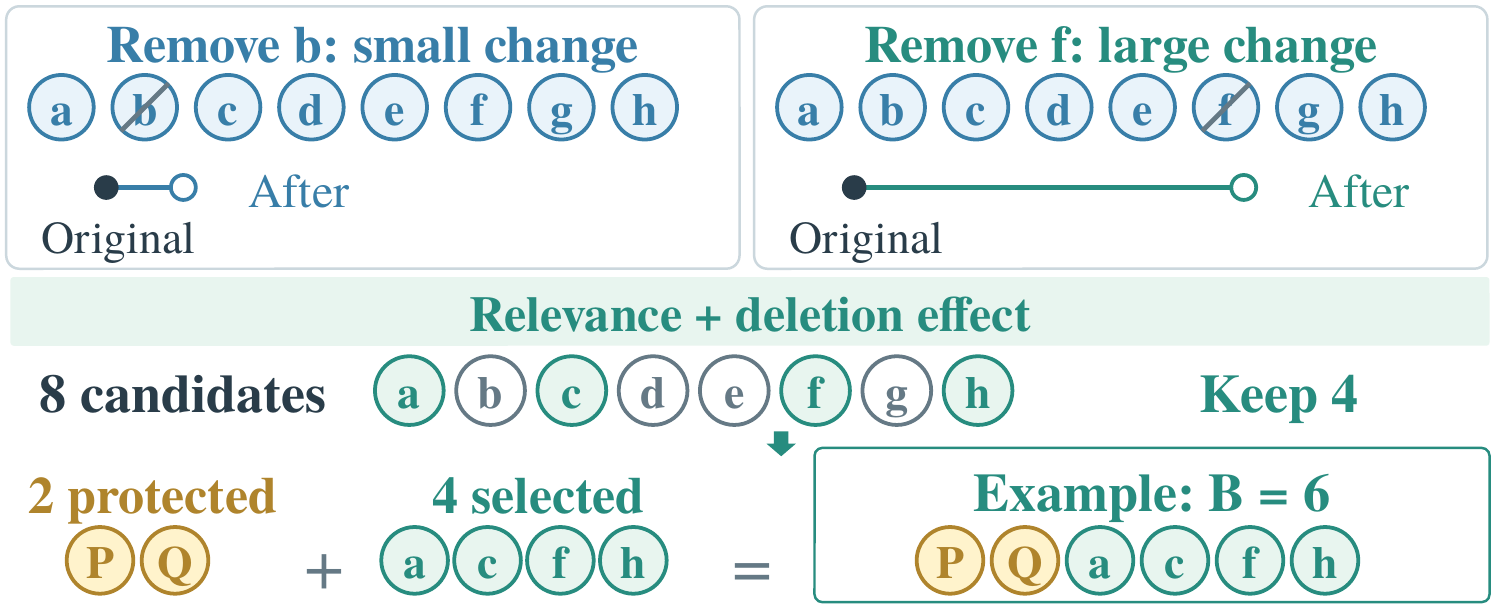}
\caption{\textbf{Filling the remaining token budget.} A larger deletion-induced output change signals greater influence. Joint ranking combines this effect with attention importance and key redundancy. In this schematic example, two protected tokens and four of eight candidates fill budget $B=6$.}
\label{fig:refinement_illustration}
\end{figure}

\paragraph{Relevance and redundancy scores.}
First, we aggregate recent-query attention to measure each token's relevance to current reasoning, prioritizing tokens receiving more attention. Second, we estimate redundancy from similarities between cached keys to reduce repeated information within the limited cache. R-KV combines these signals to favor relevant, less redundant tokens. We denote its score by $b_i$ and fuse it with the singleton deletion effect below.

\paragraph{Deletion effect.}
We also assess how much removing a token changes the attention output. Applying the definition in Section~\ref{sec:Problem_Formulation} to token $i$ gives
\begin{equation}
 s_i=\mathcal R(\{i\};\mathcal Q),
\label{eq:token_deletion_effect}
\end{equation}
where $\mathcal Q$ is the recent query window. Larger effects indicate greater output changes and higher retention priority~\citep{guo2024vatp,geng2025accurate,chang2026value}. The paired analysis in Section~\ref{sec:observations} finds that this additional signal reduces output error on held-out subsequent queries at matched budgets.

\begin{table*}[t!]
\centering
\captionsetup{skip=2pt}
\fontsize{8}{8.3}\selectfont
\setlength{\tabcolsep}{1.5pt}
\renewcommand{\arraystretch}{0.98}
\setlength{\aboverulesep}{1pt}
\setlength{\belowrulesep}{1.2pt}
\caption{\textbf{Results across eight benchmarks for four multimodal models at a 512-token KV-cache budget.} Scores are percentages; Average is the unweighted mean across benchmarks. Bold marks the best compressed results, including ties.}
\label{tab:main_results}
\begin{tabularx}{\textwidth}{@{}>{\raggedright\arraybackslash}p{0.23\textwidth}K{1}K{2}K{3}K{4}K{5}K{6}K{7}K{8}K{9}@{}}
\toprule
& \multicolumn{4}{c}{\textbf{Tool/evidence}} & \multicolumn{4}{c}{\textbf{Visual/reasoning}} & \textbf{Overall} \\
\cmidrule(lr){2-5}\cmidrule(lr){6-9}\cmidrule(l){10-10}
\textbf{Method} & \textbf{MMSearch} & \textbf{A-OKVQA} & \textbf{ViQuAE} & \textbf{V*Bench} & \textbf{RWQA} & \textbf{MathVista} & \textbf{MMVP} & \textbf{OCRBench} & \textbf{Average} \\
\midrule

\multicolumn{10}{c}{\cellcolor{kvsoftblue!45}\textbf{Qwen3-VL-8B-Instruct}}\\
\midrule
\noalign{\kvbestrow{000000000}}
\rowcolor{kvsoftgray}FullKV & 35.00 & 87.16 & 55.61 & 89.01 & 73.20 & 78.10 & 61.33 & 86.70 & 70.76 \\
\noalign{\kvbestrow{000000100}}
SnapKV~\cite{snapkv} (NeurIPS'24) & 29.67 & 82.79 & 40.65 & 75.92 & 67.06 & 72.70 & 61.33 & 83.60 & 64.22 \\
\noalign{\kvbestrow{000000000}}
R-KV~\cite{rkv} (NeurIPS'25) & 30.00 & 83.06 & 52.74 & 86.91 & 65.75 & 73.00 & 60.67 & 83.70 & 66.98 \\
\noalign{\kvbestrow{000000000}}
TriAttention~\cite{triattention} (ICML'26) & 23.67 & 86.03 & 47.81 & 83.25 & 69.28 & 72.50 & 58.67 & 70.00 & 63.90 \\
\noalign{\kvbestrow{001001000}}
AgentKV~\cite{agentkv2026} (EMNLP'26) & 30.33 & 86.64 & 57.60 & 82.72 & 69.80 & 74.10 & 60.00 & 83.40 & 68.07 \\
\noalign{\kvbestrow{111110011}}
\rowcolor{kvsoftcyan}\textbf{\method{}} & 34.67 & 87.07 & 57.60 & 90.58 & 69.93 & 74.00 & 60.67 & 84.30 & 69.85 \\

\midrule
\multicolumn{10}{c}{\cellcolor{kvsoftblue!45}\textbf{Qwen3.8-27B}}\\
\midrule
\noalign{\kvbestrow{000000000}}
\rowcolor{kvsoftgray}FullKV & 69.33 & 92.05 & 67.62 & 96.34 & 87.71 & 88.40 & 73.33 & 83.30 & 82.26 \\
\noalign{\kvbestrow{000000000}}
SnapKV~\cite{snapkv} (NeurIPS'24) & 41.67 & 69.08 & 38.27 & 29.32 & 52.42 & 63.70 & 38.00 & 72.70 & 50.64 \\
\noalign{\kvbestrow{000001000}}
R-KV~\cite{rkv} (NeurIPS'25) & 42.33 & 76.77 & 51.71 & 32.98 & 61.96 & 64.60 & 57.33 & 71.90 & 57.45 \\
\noalign{\kvbestrow{000000000}}
TriAttention~\cite{triattention} (ICML'26) & 30.33 & 78.34 & 35.64 & 36.13 & 58.56 & 51.20 & 60.67 & 52.80 & 50.46 \\
\noalign{\kvbestrow{000000000}}
AgentKV~\cite{agentkv2026} (EMNLP'26) & 34.33 & 72.75 & 47.81 & 36.65 & 63.79 & 52.50 & 60.67 & 67.00 & 54.44 \\
\noalign{\kvbestrow{111111111}}
\rowcolor{kvsoftcyan}\textbf{\method{}} & 42.67 & 82.79 & 62.29 & 65.45 & 70.46 & 64.60 & 69.33 & 75.50 & 66.64 \\

\midrule
\multicolumn{10}{c}{\cellcolor{kvsoftblue!45}\textbf{Phi-4-Reasoning-Vision-15B}}\\
\midrule
\noalign{\kvbestrow{000000000}}
\rowcolor{kvsoftgray}FullKV & 43.33 & 86.11 & 47.18 & 63.35 & 69.02 & 73.90 & 50.00 & 75.40 & 63.54 \\
\noalign{\kvbestrow{000000000}}
SnapKV~\cite{snapkv} (NeurIPS'24) & 30.67 & 82.45 & 47.81 & 63.35 & 68.89 & 69.80 & 43.33 & 68.30 & 59.33 \\
\noalign{\kvbestrow{100000000}}
R-KV~\cite{rkv} (NeurIPS'25) & 37.67 & 80.17 & 48.21 & 62.83 & 68.76 & 72.50 & 40.67 & 71.30 & 60.26 \\
\noalign{\kvbestrow{000000100}}
TriAttention~\cite{triattention} (ICML'26) & 26.33 & 65.68 & 37.71 & 63.35 & 68.89 & 71.30 & 48.67 & 60.30 & 55.28 \\
\noalign{\kvbestrow{000000000}}
AgentKV~\cite{agentkv2026} (EMNLP'26) & 31.33 & 83.14 & 43.68 & 63.35 & 69.02 & 69.70 & 41.33 & 60.60 & 57.77 \\
\noalign{\kvbestrow{111111011}}
\rowcolor{kvsoftcyan}\textbf{\method{}} & 37.67 & 83.84 & 48.69 & 67.02 & 69.15 & 72.90 & 44.67 & 72.20 & 62.02 \\

\midrule
\multicolumn{10}{c}{\cellcolor{kvsoftblue!45}\textbf{MiniCPM-V-4.7-35B-A3B}}\\
\midrule
\noalign{\kvbestrow{000000000}}
\rowcolor{kvsoftgray}FullKV & 59.00 & 92.40 & 71.44 & 83.77 & 74.51 & 87.60 & 64.00 & 89.20 & 77.74 \\
\noalign{\kvbestrow{000000000}}
SnapKV~\cite{snapkv} (NeurIPS'24) & 48.33 & 84.19 & 59.35 & 80.10 & 70.72 & 80.60 & 45.33 & 84.90 & 69.19 \\
\noalign{\kvbestrow{000000000}}
R-KV~\cite{rkv} (NeurIPS'25) & 49.67 & 89.17 & 68.66 & 72.25 & 67.58 & 78.40 & 40.00 & 86.60 & 69.04 \\
\noalign{\kvbestrow{000100100}}
TriAttention~\cite{triattention} (ICML'26) & 42.67 & 89.69 & 61.89 & 82.72 & 63.79 & 74.40 & 58.00 & 64.20 & 67.17 \\
\noalign{\kvbestrow{000000000}}
AgentKV~\cite{agentkv2026} (EMNLP'26) & 50.67 & 91.09 & 70.41 & 77.49 & 72.29 & 80.80 & 55.33 & 85.40 & 72.94 \\
\noalign{\kvbestrow{111011011}}
\rowcolor{kvsoftcyan}\textbf{\method{}} & 53.33 & 91.18 & 70.88 & 80.10 & 72.55 & 80.90 & 56.67 & 88.10 & 74.21 \\
\bottomrule
\end{tabularx}
\end{table*}

\paragraph{Final score and selection.}
To align their scales, we convert $b_i$ and $s_i$ to percentile ranks $\widehat b_i$ and $\widehat s_i$ among eligible candidates, with larger ranks indicating higher priority. Fusion uses relative candidate positions, avoiding direct comparison of raw scores with different scales. The final score is
\begin{equation}
 f_i=(1-\lambda)\widehat b_i+\lambda\widehat s_i,
\label{eq:fused_token_score}
\end{equation}
where $\lambda$ weights the deletion-effect rank. Increasing $\lambda$ gives deletion effects more influence over the ordering, while $\lambda=0$ recovers relevance-based ranking. The highest-scoring candidates fill the remaining slots and join $\mathcal N_j$ to form $\mathcal I_j$. For example, with $B=512$ and 128 protected tokens, we select the top 384 other candidates. Scoring runs independently per layer and KV head, reusing existing attention quantities without extra forward passes. Implementation details appear in Appendices~\ref{sec:retained_index_construction} and~\ref{sec:visual_refinement_dev}; Figure~\ref{fig:refinement_illustration} illustrates the allocation.

Algorithm~\ref{alg:commitkv} summarizes the procedure.

\begin{algorithm}[tb]
\caption{\method{} compression procedure}
\label{alg:commitkv}
\small
\begin{algorithmic}[1]
\Require User query, frozen VLM, tools, budget $B$
\Ensure Answer using the compressed KV cache
\State Initialize the cache and page records
\While{the final answer has not been generated}
  \State Advance native generation or tool use
  \State Update available measurements and page lifecycles (Secs.~\ref{sec:commit}--\ref{sec:transition})
  \If{compression is due}
    \State Compress within $B$ using mandatory tokens and fused scores (Sec.~\ref{sec:visual_refinement})
  \EndIf
\EndWhile
\State \Return the final answer
\end{algorithmic}
\end{algorithm}

\section{Experiments}
\label{sec:experiments}

\subsection{Experimental Setup}
\label{sec:repeat_protocol}
\label{sec:experimental_setup}

\paragraph{Models.}
We evaluate Qwen3-VL-8B-Instruct~\citep{qwen3vl_8b_instruct}, Qwen3.8-27B~\citep{qwen38_27b}, Phi-4-Reasoning-Vision-15B~\citep{phi4_reasoning_vision_15b}, and MiniCPM-V-4.7-35B-A3B~\citep{minicpmv47_35b_a3b}. These open-weight models span 8B--35B parameters, dense and mixture-of-experts (MoE) architectures, and hybrid linear/full attention (Appendix~\ref{sec:Details_of_Models}).

\paragraph{Baselines.}
We compare FullKV and four compression methods: SnapKV~\citep{snapkv}, R-KV~\citep{rkv}, TriAttention~\citep{triattention}, and AgentKV~\citep{agentkv2026}. FullKV retains the complete cache. The first three compression methods use attention or relevance scores; AgentKV selects phase-aware queries (Appendix~\ref{sec:Details_of_Baselines}).

\paragraph{Benchmarks.}
Eight benchmarks assess two capabilities. MMSearch~\citep{mmsearch2024}, A-OKVQA~\citep{aokvqa2022}, ViQuAE~\citep{viquae2022}, and V*Bench~\citep{vstar2023} evaluate evidence acquisition and use through retrieval and visual search. RealWorldQA~\citep{realworldqa2024}, MathVista~\citep{mathvista2023}, MMVP~\citep{mmvp2024}, and OCRBench v1~\citep{ocrbench2024} evaluate spatial understanding, visual mathematics, fine-grained discrimination, and text recognition. We report benchmark scores and their unweighted average. Details and tool-use statistics appear in Appendix~\ref{sec:Details_of_Benchmarks} and Table~\ref{tab:trajectory_stats}.

\paragraph{Implementation Details.}
We implement \method{} with SGLang~\citep{sglang}, vLLM~\citep{vllm}, and Transformers on NVIDIA H20 96GB GPUs. Defaults are $W=8$, $G=16$, $\tau_{\mathrm{use}}=0.1$, $\tau_{\mathrm{dead}}=0.01$, and $\lambda=0.25$. Compression occurs every 128 decoded tokens, retaining at most $B\in\{256,512,1024\}$ tokens after each checkpoint. Serving and evaluation protocols are in Appendix~\ref{sec:experimental_details}.

\subsection{Main Results}
\label{sec:main_results}

\paragraph{Comparison with State-of-the-Art Methods.}
Table~\ref{tab:main_results} compares eight-benchmark accuracy at a 512-token budget. \method{} leads compression methods in average accuracy on all four models, with absolute gains of 1.78\%, 9.19\%, 1.76\%, and 1.27\% over the strongest baseline on Qwen3-VL, Qwen3.8, Phi, and MiniCPM, respectively. Qwen3-VL achieves 69.85\% versus FullKV's 70.76\%, a loss of only 0.91\%. These results demonstrate the effectiveness and superiority of \method{} across multimodal architectures.

\begin{figure}[t]
\centering
\includegraphics[width=\columnwidth]{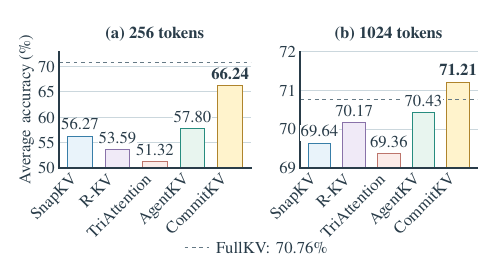}
\captionsetup{skip=2pt}
\caption{\textbf{Average accuracy at 256 and 1024 tokens (Qwen3-VL-8B).} Eight-benchmark means; dashed lines denote FullKV. Full per-dataset results are in Appendix Table~\ref{tab:different_budgets_results}.}
\label{fig:cache_budgets}
\end{figure}

\paragraph{Results at Different Cache Budgets.}
\method{} also leads the compared compression methods across Qwen3-VL-8B-Instruct's three budgets (Table~\ref{tab:main_results}, Figure~\ref{fig:cache_budgets}). At 256, 512, and 1024 tokens, average accuracies are 66.24\%, 69.85\%, and 71.21\%, exceeding the strongest baseline by 8.44\%, 1.78\%, and 0.78\%. Compared with FullKV, accuracy decreases by 4.52\% and 0.91\% at 256 and 512 tokens, and increases by 0.45\% at 1024. The larger advantage under the most aggressive budget and consistent lead across budgets demonstrate robustness to cache size.

\subsection{Efficiency Analyses}
\label{sec:efficiency_analyses}
\label{sec:memory_latency}

\begin{figure}[!t]
\centering
\includegraphics[width=\columnwidth]{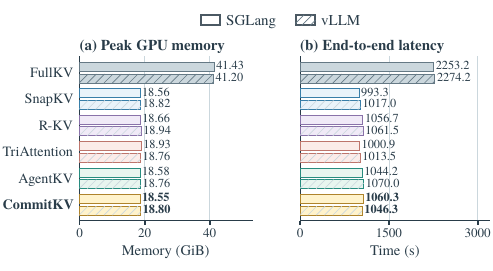}
\captionsetup{skip=2pt}
\caption{\textbf{Efficiency at a 256-token KV-cache budget.} Peak allocated GPU memory and end-to-end latency with 8,192 input and 163,840 generated tokens. Solid bars denote SGLang and hatched bars vLLM; gray denotes each backend's FullKV reference. Lower is better. Complete results are in Appendix~\ref{app:efficiency_protocols}.}
\label{fig:memory_latency}
\end{figure}

\begin{figure*}[t]
\centering
\includegraphics[width=\textwidth]{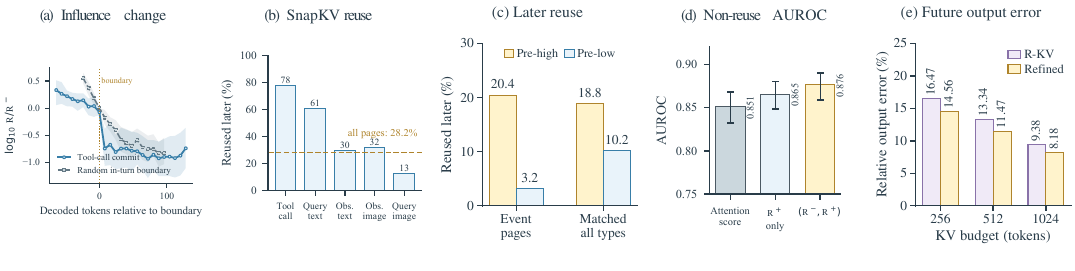}
\caption{\textbf{Further analysis of the three design choices.} (a) Influence around tool versus random boundaries (mean/interquartile range). (b) Later reuse under SnapKV scoring (Q/O: query/observation; t/i: text/image). (c) Reuse for post-low tool-call/observation pages and post-effect-matched groups across all page types. (d) Non-reuse AUROC with 95\% trajectory-bootstrap intervals. (e) Subsequent-query output error with/without deletion-effect refinement at equal budgets. Panels (a)--(d) use 60 FullKV trajectories; (e) uses 32 paired probes from 16 trajectories.}
\label{fig:observation}
\end{figure*}

We benchmark Qwen3-VL-8B-Instruct on NVIDIA H20 96GB GPUs using SGLang and vLLM, comparing FullKV, four compression baselines, and \method{}. At $B=256$, all methods greedily generate 163,840 tokens from the same 8,192-token input (Figure~\ref{fig:memory_latency}). Complete comparisons, other budgets, and generation lengths are in Appendix~\ref{app:efficiency_protocols}.

Relative to each backend's FullKV, \method{} achieves $2.12\times$ and $2.17\times$ end-to-end speedups in SGLang and vLLM. Peak allocated GPU memory is 18.55 and 18.80 GiB, reducing memory by 55.23\% and 54.37\%, respectively. Both backends benefit in latency and memory, supporting efficient deployment.

\subsection{Ablation Studies}
\label{sec:ablation_studies}

\paragraph{Component Ablations.}
Table~\ref{tab:ablation_studies} tests the components on Qwen3-VL-8B-Instruct at $B=256$, reporting MMSearch, V*Bench, and MMVP accuracy and their unweighted average. The complete method achieves 57.92\%. We first remove stale-page retirement, allowing those pages to remain candidates for selection; the average falls to 56.73\%. We then remove recently used page protection, so these pages compete with the other candidates, obtaining 57.02\%. Finally, removing deletion-effect refinement while retaining lifecycle decisions yields 51.23\%. Each experiment removes only the specified component and retains the other decisions. All three removals reduce average accuracy, supporting the combined page-management and token-selection design.

\begin{table}[!ht]
\centering\footnotesize
\setlength{\tabcolsep}{2.1pt}
\renewcommand{\arraystretch}{1.03}
\caption{\textbf{Single-component removals at a 256-token budget on Qwen3-VL-8B-Instruct.} Bold: column maxima. Full references, removal configurations, and the joint-removal control are detailed in Appendix~\ref{app:component_ablation_protocol}.}
\label{tab:ablation_studies}
\begin{tabular}{@{}lrrrr@{}}
\toprule
Variant & MMSearch & V*Bench & MMVP & Avg. \\
\midrule
\rowcolor{kvsoftcyan}
\textbf{Full \method{}} & \textbf{30.00\%} & \textbf{83.77\%} & \textbf{60.00\%} & \textbf{57.92\%} \\
\midrule
w/o Retirement & 29.33\% & 82.20\% & 58.67\% & 56.73\% \\
w/o Protection & 29.67\% & 82.72\% & 58.67\% & 57.02\% \\
w/o Deletion Refinement & 22.00\% & 74.35\% & 57.33\% & 51.23\% \\
\bottomrule
\end{tabular}
\end{table}

\begin{figure}[!ht]
\centering
\includegraphics[width=\columnwidth]{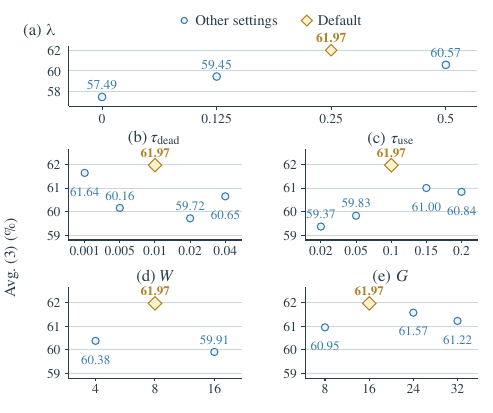}
\captionsetup{skip=2pt}
\caption{\textbf{Hyperparameter sensitivity at a 512-token KV-cache budget.} Three-benchmark averages on MMSearch, V*Bench, and MMVP. Blue circles denote other settings; gold diamonds mark defaults.}
\label{fig:hyperparameter_main}
\end{figure}

\paragraph{Hyperparameter Sensitivity.}
Figure~\ref{fig:hyperparameter_main} varies one parameter at a time across 16 non-default settings on Qwen3-VL-8B-Instruct at $B=512$, reporting the MMSearch/V*Bench/MMVP average. The defaults $\lambda=0.25$, $\tau_{\mathrm{dead}}=0.01$, $\tau_{\mathrm{use}}=0.1$, $W=8$, and $G=16$ achieve 61.97\%. For $\lambda=0,0.125,0.5$, averages are 57.49\%, 59.45\%, and 60.57\%; the default gains 4.48\% over $\lambda=0$, supporting the deletion-effect contribution to joint scoring. Varying $\tau_{\mathrm{dead}}$ or $\tau_{\mathrm{use}}$ gives 59.72\%--61.64\% or 59.37\%--61.00\%; varying $W$ or $G$ gives 59.91\%--60.38\% or 60.95\%--61.57\%. Defaults yield the highest tested average. Complete results appear in Appendix~\ref{app:task_hyperparameter_sensitivity}.

\subsection{Further Analysis}
\label{sec:observations}
We revisit Section~\ref{sec:motivation} through measurement timing, page reuse, and joint scoring (Figure~\ref{fig:observation}). Diagnostics use recorded states at Qwen3-VL-8B-Instruct's last full-attention layer: 60 FullKV trajectories for the first two analyses and 32 paired probes from 16 trajectories for scoring.

\phantomsection\label{obs:commit}
\paragraph{Commit-aligned measurement.}
We compare page influence around tool-call and random within-turn boundaries using equal decoding windows. Across 494 call-page measurements, influence falls by $5.6\times$ on average within eight decoded tokens after the observation, versus $1.5\times$ after random boundaries. Random-boundary pages need about 48 tokens for a comparable decline (Figure~\ref{fig:observation}a). These sharper changes support measuring both sides of tool interactions to capture a page's earlier role (Section~\ref{sec:commit}).

\phantomsection\label{obs:transition}
\paragraph{Lifecycle-aware compression.}
In FullKV replay, 28.2\% of pages selected for eviction by SnapKV scores are reused later (Figure~\ref{fig:observation}b). To examine what pre-call influence adds, we analyze 7{,}886 tool-call and observation-page measurements. Among post-low pages, pre-high pages have 20.4\% later reuse, versus 3.2\% for pages low on both sides. Matching post-observation influence across all page types gives 18.8\% versus 10.2\% reuse for pre-high and pre-low groups (Figure~\ref{fig:observation}c). Paired effects predict non-reuse with AUROC 0.876, versus 0.865 for post-only effects and 0.851 for attention (Figure~\ref{fig:observation}d). Pre-call influence thus provides information beyond the current score, supporting retirement when both effects are low and temporary protection for recently used pages (Section~\ref{sec:transition}). Group definitions and controls appear in Appendices~\ref{sec:extended_observations}, \ref{sec:matched_transition_validation}, and~\ref{app:new_transitions}.

\phantomsection\label{obs:refinement}
\paragraph{Joint scoring.}
We compare relevance-only selection with joint scoring on 32 paired probes, holding the token budget, recent tokens, and evaluation queries fixed. Eight queries determine the retained set, and the next eight evaluate output error relative to FullKV; page-level reservations are disabled to isolate scoring. At 256, 512, and 1024 tokens, adding singleton deletion effects reduces mean output error from 16.47\%, 13.34\%, and 9.38\% to 14.56\%, 11.47\%, and 8.18\%, respectively (Figure~\ref{fig:observation}e). All 32 pairs improve at every budget. Joint scoring thus better preserves subsequent attention outputs at equal budgets, supporting Section~\ref{sec:visual_refinement}. Settings are in Appendix~\ref{app:frozen_fidelity}.

\section{Conclusion}
\label{sec:conclusion}

This paper presents \method{}, a training-free KV cache compression method for multimodal agents. By combining commit-aligned influence measurement, page retirement and temporary protection, and token-level deletion-effect refinement, it preserves useful evidence within a fixed cache budget. Experiments across four models and eight benchmarks demonstrate higher average accuracy than the compared compression methods, together with lower GPU memory use and faster inference than FullKV, supporting efficient multimodal agent deployment.

\FloatBarrier
\section*{Generative AI Disclosure}
GPT and Codex assisted with literature review, mathematical derivations, manuscript drafting and language editing, figure preparation, and code development and debugging. The authors independently reviewed and validated the outputs and take full responsibility for this work.

\clearpage
\appendix
\section*{Appendix}
\pdfbookmark[0]{Appendix}{appendix_overview}
\makeatletter
\raggedbottom
\setcounter{dbltopnumber}{2}\setcounter{topnumber}{2}\setcounter{totalnumber}{3}
\renewcommand{\dbltopfraction}{.85}\renewcommand{\topfraction}{.75}
\renewcommand{\textfraction}{.10}\renewcommand{\floatpagefraction}{.65}\renewcommand{\dblfloatpagefraction}{.65}
\setlength{\textfloatsep}{6pt plus 1pt minus 1pt}\setlength{\dbltextfloatsep}{7pt plus 1pt minus 1pt}
\setlength{\floatsep}{6pt plus 1pt minus 1pt}\setlength{\dblfloatsep}{9pt plus 1pt minus 1pt}
\setlength{\@fptop}{0pt}\setlength{\@dblfptop}{0pt}
\setlength{\@fpsep}{7pt plus 1pt minus 1pt}\setlength{\@dblfpsep}{9pt plus 1pt minus 1pt}
\setlength{\@fpbot}{0pt plus 1fil}\setlength{\@dblfpbot}{0pt plus 1fil}
\setlength{\intextsep}{7pt plus 1pt minus 1pt}
\setlength{\abovedisplayskip}{6pt plus 1pt minus 1pt}\setlength{\belowdisplayskip}{6pt plus 1pt minus 1pt}
\setlength{\abovedisplayshortskip}{3pt plus 1pt minus 1pt}\setlength{\belowdisplayshortskip}{4pt plus 1pt minus 1pt}
\captionsetup{skip=3pt}
\makeatother

\section{Evaluation Protocol}
\label{sec:experimental_details}

\begin{samepage}
\subsection{Agent Tools and ReAct Trajectories}
\label{app:agent_tools}

MMSearch, A-OKVQA, and ViQuAE use web search (Google via Serper) for titles and snippets, text-to-image search for images and sources, reverse image search (Google Lens) for matches to an image or crop, and browser visits for page text and screenshots. Crop and zoom returns a region specified by an image ID and bounding box. Python runs short scripts and returns text output. MMVP and OCRBench v1 offer cropping alone. The other tasks also offer Python. Tool outputs are appended to the dialogue as observations.

Table~\ref{tab:trajectory_stats} summarizes tool use under FullKV.

\nopagebreak[4]
\begin{table}[H]
\centering
\captionsetup{skip=2pt}
\footnotesize
\setlength{\tabcolsep}{2.5pt}
\caption{\textbf{Qwen3-VL-8B-Instruct trajectories under FullKV.}}
\label{tab:trajectory_stats}
\resizebox{\linewidth}{!}{
\begin{tabular}{@{}lcc|ccccc@{}}
\toprule
& \multicolumn{2}{c|}{\textbf{All trajectories}} & \multicolumn{5}{c}{\textbf{Selected trajectories}} \\
\textbf{Benchmark} & $\boldsymbol{\geq}$\textbf{1 call} & $\boldsymbol{\geq}$\textbf{2 calls} & $n$ & \textbf{Turns} & \textbf{Calls} & \textbf{Vis.\ (\%)} & \textbf{Length} \\
\midrule
MMSearch & 67.7 & 18.7 & 15 & 3.53 & 2.53 & 20.5 & 7{,}228 \\
A-OKVQA & 59.9 & 10.9 & 15 & 3.47 & 2.47 & 32.8 & 2{,}607 \\
ViQuAE & 87.7 & 22.2 & 270 & 3.70 & 2.70 & 23.0 & 8{,}200 \\
V*Bench & 97.9 & 23.6 & 20 & 3.05 & 2.05 & 75.7 & 5{,}092 \\
RealWorldQA & 96.7 & 40.7 & 10 & 3.10 & 2.10 & 59.0 & 3{,}240 \\
OCRBench v1 & 26.7 & 4.1 & 41 & 3.51 & 2.51 & 58.2 & 3{,}338 \\
\bottomrule
\end{tabular}}
\end{table}
\end{samepage}

\subsection{Main-Text Component Ablation Protocol}
\label{app:component_ablation_protocol}

At $B=256$, each single-removal run changes only its named component. Table~\ref{tab:component_ablation_reference} gives the full references and the joint removal of retirement and protection.

\begin{table}[H]
\centering\footnotesize
\setlength{\tabcolsep}{3.2pt}
\caption{\textbf{Full reference and joint-removal scores at $B=256$.} Scores are percentages.}
\label{tab:component_ablation_reference}
\begin{tabular*}{\linewidth}{@{\extracolsep{\fill}}lrrrr@{}}
\toprule
Setting & MMSearch & V*Bench & MMVP & Mean\\
\midrule
Full reference & 30.00 & 83.77 & 60.00 & 57.92\\
Joint removal & 27.33 & 82.20 & 58.00 & 55.84\\
\bottomrule
\end{tabular*}
\end{table}

\subsection{Serving Settings}
\label{app:serving_settings}

Task runs serve both Qwen models with SGLang and Phi and MiniCPM-V with Transformers. Qwen3-VL re-prefills the accumulated dialogue at each request with radix reuse disabled. Standard temperatures are 0.7 for Qwen3-VL and MiniCPM-V and 0 for Phi. Qwen3.8 uses its released settings. Per-request output/context caps are 16,384/65,536 for Qwen3-VL and MiniCPM-V, 32,768/65,536 for Qwen3.8, and 4,096/16,384 for Phi. For Qwen3-VL on MathVista the caps are 32,768/128,000. The efficiency runtimes are described in Appendix~\ref{app:efficiency_protocols}.

\section{Theoretical Analysis of Commit Transitions}
\label{sec:theory}

We derive a closed form for the page-deletion effect defined in Section~\ref{sec:Problem_Formulation}, Eq.~\eqref{eq:window_residual}. It shows how to compute the attention-output change from existing attention weights and value vectors without an extra forward pass.

\begin{samepage}

\subsection{Exact Form of the Page-Deletion Effect}
\label{app:exact_page_deletion}

For one head and query $p$, let $a_{pi}$ be normalized causal attention weights, $\boldsymbol v_i$ the value vector of token $i$, and $\boldsymbol o_p$ the original attention output. A deletable page $\mathcal E$ has mass and value sum
\begin{equation}
m_{\mathcal E}
=
\sum_{\substack{i\in \mathcal E\\p_i\leq p}}a_{pi},
\qquad
\boldsymbol s_{\mathcal E}
=
\sum_{\substack{i\in \mathcal E\\p_i\leq p}}
a_{pi}\boldsymbol v_i .
\label{eq:theory_mass_sum}
\end{equation}
\end{samepage}
For $m_{\mathcal E}>0$, its mean value is $\bar{\boldsymbol v}_{\mathcal E}=\boldsymbol s_{\mathcal E}/m_{\mathcal E}$.

Removing page $\mathcal E$ and renormalizing the remaining attention gives the closed form
\begin{equation}
\widetilde{\boldsymbol o}_p
=\frac{\boldsymbol o_p-\boldsymbol s_{\mathcal E}}{1-m_{\mathcal E}}.
\label{eq:closed_residual}
\end{equation}

\paragraph{Proposition 1 (deletion-effect decomposition).}
If $m_{\mathcal E}<1$, deleting $\mathcal E$ and renormalizing the remaining attention gives
\begin{equation}
\boldsymbol o_p-\widetilde{\boldsymbol o}_p
=
\frac{\boldsymbol s_{\mathcal E}-m_{\mathcal E}\boldsymbol o_p}{1-m_{\mathcal E}}
=
\frac{m_{\mathcal E}}{1-m_{\mathcal E}}
\left(
\bar{\boldsymbol v}_{\mathcal E}-\boldsymbol o_p
\right),
\label{eq:deletion_decomposition}
\end{equation}
where the second equality applies when $m_{\mathcal E}>0$. Consequently, for
$0<m_{\mathcal E}<1$ and $\|\boldsymbol o_p\|_2>0$,
\begin{equation}
\frac{
\|\boldsymbol o_p-\widetilde{\boldsymbol o}_p\|_2
}{
\|\boldsymbol o_p\|_2
}
=
\frac{m_{\mathcal E}}{1-m_{\mathcal E}}
\frac{
\|\bar{\boldsymbol v}_{\mathcal E}-\boldsymbol o_p\|_2
}{
\|\boldsymbol o_p\|_2
}.
\label{eq:deletion_factorization}
\end{equation}

\emph{Proof.} Subtract Eq.~\eqref{eq:closed_residual} from $\boldsymbol o_p$ to obtain Eq.~\eqref{eq:deletion_decomposition}. Substitute $\boldsymbol s_{\mathcal E}=m_{\mathcal E}\bar{\boldsymbol v}_{\mathcal E}$ and take norms to obtain Eq.~\eqref{eq:deletion_factorization}. The effect is zero when $m_{\mathcal E}=0$.
\hfill$\square$

Equation~\eqref{eq:deletion_factorization} separates attention mass from value contrast.

\section{Implementation Details}
\label{app:algorithm_implementation}

\subsection{Construction of the Retained-Token Index Set}
\label{sec:retained_index_construction}

Sections~\ref{sec:transition} and~\ref{sec:visual_refinement} define retirement, protection, and joint scoring. We detail how these decisions produce the retained row indices at each compression checkpoint.

At checkpoint $j$, resident row $i$ of layer $\ell$ and KV head $h$ has absolute position $p_{j,i}$. We omit $(\ell,h)$ from the row sets below. Page labels use absolute positions and apply to the corresponding rows of each head.

\begin{align}
 \mathcal J_j&=\{i:p_{j,i}\in \mathcal E\text{ for some }\mathcal E\in\mathcal D\},
 \label{eq:retired_token_indices}\\
 \mathcal C_j&=\{1,\ldots,S_j\}\setminus\mathcal J_j.
 \label{eq:survivor_pool}
\end{align}

Among surviving rows, reserve complete anchors first, then protect recently used pages (Eq.~\eqref{eq:protect_rule}) in decreasing order of $\mathcal R_c^{-}$. Together they form $\mathcal N_j$ and use at most $\lfloor\rho B\rfloor$ slots ($\rho=0.25$ by default), counting overlaps once. Pages that do not fit as a whole remain eligible for ordinary selection.

\begin{align}
 b_j&=\min\{B,|\mathcal C_j|\}-|\mathcal N_j|,
 \label{eq:ordinary_selection_budget}\\
 \mathcal I_j&=\mathcal N_j\cup\mathcal O_j,\quad
 \mathcal O_j\subseteq\mathcal C_j\setminus\mathcal N_j,\quad |\mathcal O_j|=b_j.
 \label{eq:retained_set_construction}
\end{align}

Select the $b_j$ rows with the highest fused scores $f_i$ (Eq.~\eqref{eq:fused_token_score}) from $\mathcal C_j\setminus\mathcal N_j$. Compute percentile ranks within these candidates for each layer and KV head, using midranks for ties. Section~\ref{sec:experimental_setup} gives the default $\lambda$. When $|\mathcal C_j|\le B$, all surviving rows are retained.

To compute the singleton effect in Eq.~\eqref{eq:token_deletion_effect}, use the latest $W$ queries in $\mathcal Q_j$. With attention weight $a_{pi}$, value $\boldsymbol v_i$, and full output $\boldsymbol o_p$, the stabilized effect is

\begin{equation*}
\begin{aligned}
\delta_{pi}
&=\frac{a_{pi}}{\max(1-a_{pi},\epsilon)}
  \frac{\|\boldsymbol v_i-\boldsymbol o_p\|_2}
       {\max(\|\boldsymbol o_p\|_2,\epsilon)},\\
s_i^{(\ell,h)}
&=\max_{r\in\mathcal H(h)}\max_{p\in\mathcal Q_j}
  \delta_{pi}^{(\ell,r)},
\end{aligned}
\end{equation*}

Here $\epsilon=10^{-6}$ and $\mathcal H(h)$ is the set of query heads sharing KV head $h$. Each layer and KV head selects its own $\mathcal I_j$ and uses those same indices to gather its keys, values, and absolute positions.

\section{Joint Scoring Details}
\label{sec:standing_details}

\subsection{Joint Scoring Implementation}
\label{sec:visual_refinement_dev}

At a checkpoint that needs token selection, we group the cached queries in $\mathcal Q_j$ by their shared KV head and compute causal attention over the resident cache. To evaluate the singleton deletion effect in Appendix~\ref{sec:retained_index_construction}, we compute the value--output distances in a batch using
\begin{equation*}
\|\boldsymbol v_i-\boldsymbol o_p\|_2^2
=\|\boldsymbol v_i\|_2^2+\|\boldsymbol o_p\|_2^2
 -2\boldsymbol v_i^\top\boldsymbol o_p .
\end{equation*}
We clamp the squared distance to zero if rounding makes it negative, take its square root, and reduce the resulting deletion effects by the maximum over queries and query heads sharing the same KV head. For each score, we normalize its zero-based midrank by $n-1$ among the $n$ eligible text and visual tokens, using zero when $n=1$. We fuse the two ranks with Eq.~\eqref{eq:fused_token_score} and use the remaining budget from Appendix~\ref{sec:retained_index_construction}.

\section{Models, Baselines, and Benchmarks}
\label{sec:Details_of_Models_Baselines_Benchmarks}

\begin{samepage}
\subsection{Details of Models}
\label{sec:Details_of_Models}

\nopagebreak[4]
\begin{table}[H]
\centering\footnotesize
\setlength{\tabcolsep}{3pt}
\renewcommand{\arraystretch}{1.1}
\caption{\textbf{Evaluated checkpoints and architectures.}}
\label{tab:appendix_models}
\begin{tabularx}{\linewidth}{@{}p{.39\linewidth}X@{}}
\toprule
\textbf{Checkpoint} & \textbf{Architecture}\\
\midrule
\href{https://huggingface.co/Qwen/Qwen3-VL-8B-Instruct}{Qwen3-VL-8B-Instruct}~\citep{qwen3vl_8b_instruct} & Dense\\
\href{https://huggingface.co/Qwen/Qwen3.8-27B}{Qwen3.8-27B}~\citep{qwen38_27b} & Dense; hybrid linear/full attention\\
\href{https://huggingface.co/microsoft/Phi-4-reasoning-vision-15B}{Phi-4-Reasoning-Vision-15B}~\citep{phi4_reasoning_vision_15b} & Dense\\
\href{https://huggingface.co/openbmb/MiniCPM-V-4.7-35B-A3B}{MiniCPM-V-4.7-35B-A3B}~\citep{minicpmv47_35b_a3b} & MoE; about 3B active parameters\\
\bottomrule
\end{tabularx}
\end{table}
\end{samepage}

\begin{samepage}
\subsection{Details of Baselines}
\label{sec:Details_of_Baselines}

\nopagebreak[4]
\begin{table}[H]
\centering\footnotesize
\setlength{\tabcolsep}{3pt}
\renewcommand{\arraystretch}{1.1}
\caption{\textbf{Baseline retention policies.} Compression applies to full-attention layers.}
\label{tab:appendix_baselines}
\begin{tabularx}{\linewidth}{@{}lX@{}}
\toprule
\textbf{Method} & \textbf{What determines retention}\\
\midrule
FullKV & All processed token states; no eviction\\
SnapKV~\citep{snapkv} & Attention aggregated over a recent observation window; clustered selection per head\\
R-KV~\citep{rkv} & Attention importance combined with key redundancy\\
TriAttention~\citep{triattention} & Pre-RoPE query--key geometry, vector norms, and position-dependent scores\\
AgentKV~\citep{agentkv2026} & Union of representative queries from reasoning, actions, and tool observations\\
\bottomrule
\end{tabularx}
\end{table}
\end{samepage}

\begin{samepage}
\subsection{Details of Benchmarks}
\label{sec:Details_of_Benchmarks}

\nopagebreak[4]
\begin{table}[H]
\centering\footnotesize
\setlength{\tabcolsep}{2pt}
\renewcommand{\arraystretch}{1.08}
\caption{\textbf{Benchmark splits and scoring (5,958 questions per model).} All questions count. MC denotes multiple choice. MMVP uses 150 pairs. OCRBench scores are reported as percentages.}
\label{tab:benchmark_details}
\begin{tabularx}{\linewidth}{@{}p{.27\linewidth}p{.15\linewidth}rX@{}}
\toprule
\textbf{Benchmark} & \textbf{Split} & $n$ & \textbf{Score}\\
\midrule
\multicolumn{4}{c}{\textbf{Visual / reasoning}}\\
RealWorldQA~\citep{realworldqa2024} & test & 765 & MC accuracy (OpenCompass)\\
MathVista~\citep{mathvista2023} & testmini & 1000 & Accuracy; released normalization\\
MMVP~\citep{mmvp2024} & complete & 300 & Pair accuracy (both correct)\\
OCRBench v1~\citep{ocrbench2024} & complete & 1000 & Released string matching\\
\midrule
\multicolumn{4}{c}{\textbf{Tool / evidence}}\\
MMSearch~\citep{mmsearch2024} & end2end & 300 & Judge-scored answer accuracy\\
A-OKVQA~\citep{aokvqa2022} & validation & 1145 & MC accuracy\\
ViQuAE~\citep{viquae2022} & test & 1257 & Normalized exact match over aliases\\
V*Bench~\citep{vstar2023} & both subtasks & 191 & MC accuracy\\
\bottomrule
\end{tabularx}
\end{table}
\end{samepage}

\section{More Results}
\label{app:more_results}

\subsection{Results Across Cache Budgets}
\label{sec:budget_table}

Table~\ref{tab:different_budgets_results} gives the full results for Figure~\ref{fig:cache_budgets}.

\begin{table*}[!t]
\centering
\captionsetup{skip=2pt}
\footnotesize
\setlength{\tabcolsep}{1.5pt}
\renewcommand{\arraystretch}{1.00}
\caption{\textbf{Comparison of accuracy for Qwen3-VL-8B-Instruct across eight datasets with KV cache budgets of 256 and 1024 tokens.} Component ablations at 256 tokens are described in Appendix~\ref{app:component_ablation_protocol}. FullKV is a shared uncompressed reference. Average is the unweighted mean over all eight datasets. Bold marks the best compressed result for each model and budget, including ties.}
\label{tab:different_budgets_results}
\begin{tabularx}{\textwidth}{@{}>{\raggedright\arraybackslash}p{0.23\textwidth}K{1}K{2}K{3}K{4}K{5}K{6}K{7}K{8}K{9}@{}}
\toprule
& \multicolumn{4}{c}{\textbf{Tool/evidence}} & \multicolumn{4}{c}{\textbf{Visual/reasoning}} & \textbf{Overall} \\
\cmidrule(lr){2-5}\cmidrule(lr){6-9}\cmidrule(l){10-10}
\textbf{Method} & \textbf{MMSearch} & \textbf{A-OKVQA} & \textbf{ViQuAE} & \textbf{V*Bench} & \textbf{\shortstack{RealWorld\\QA}} & \textbf{MathVista} & \textbf{MMVP} & \textbf{OCRBench} & \textbf{Average} \\
\midrule
\noalign{\kvbestrow{000000000}}
\rowcolor{kvsoftgray}FullKV & 35.00 & 87.16 & 55.61 & 89.01 & 73.20 & 78.10 & 61.33 & 86.70 & 70.76 \\
\midrule
\multicolumn{10}{c}{\cellcolor{kvsoftblue!45}\textbf{Qwen3-VL-8B-Instruct --- 256-token KV budget}}\\
\midrule
\noalign{\kvbestrow{000000000}}
SnapKV~\cite{snapkv} (NeurIPS'24) & 24.00 & 77.29 & 29.83 & 64.40 & 58.56 & 66.70 & 52.67 & 76.70 & 56.27 \\
\noalign{\kvbestrow{000000000}}
R-KV~\cite{rkv} (NeurIPS'25) & 18.00 & 83.23 & 37.39 & 65.45 & 42.35 & 66.40 & 41.33 & 74.60 & 53.59 \\
\noalign{\kvbestrow{000000000}}
TriAttention~\cite{triattention} (ICML'26) & 11.00 & 74.24 & 33.49 & 58.12 & 65.36 & 65.10 & 52.67 & 50.60 & 51.32 \\
\noalign{\kvbestrow{001000000}}
AgentKV~\cite{agentkv2026} (EMNLP'26) & 21.67 & 76.16 & 54.26 & 63.87 & 48.24 & 68.20 & 58.00 & 72.00 & 57.80 \\
\noalign{\kvbestrow{110111111}}
\rowcolor{kvsoftcyan}\textbf{\method{}} & 30.00 & 83.93 & 54.02 & 83.77 & 71.11 & 69.10 & 60.00 & 78.00 & 66.24 \\
\midrule
\multicolumn{10}{c}{\cellcolor{kvsoftblue!45}\textbf{Qwen3-VL-8B-Instruct --- 1024-token KV budget}}\\
\midrule
\noalign{\kvbestrow{000010110}}
SnapKV~\cite{snapkv} (NeurIPS'24) & 34.00 & 86.29 & 52.35 & 90.05 & 71.90 & 74.50 & 61.33 & 86.70 & 69.64 \\
\noalign{\kvbestrow{000000000}}
R-KV~\cite{rkv} (NeurIPS'25) & 32.67 & 87.16 & 57.44 & 90.58 & 70.33 & 76.30 & 60.67 & 86.20 & 70.17 \\
\noalign{\kvbestrow{000100000}}
TriAttention~\cite{triattention} (ICML'26) & 31.67 & 86.03 & 55.69 & 91.10 & 70.46 & 75.80 & 60.67 & 83.50 & 69.36 \\
\noalign{\kvbestrow{000000110}}
AgentKV~\cite{agentkv2026} (EMNLP'26) & 34.67 & 86.99 & 58.15 & 87.96 & 70.85 & 76.80 & 61.33 & 86.70 & 70.43 \\
\noalign{\kvbestrow{111101111}}
\rowcolor{kvsoftcyan}\textbf{\method{}} & 36.00 & 87.60 & 58.31 & 91.10 & 71.76 & 76.90 & 61.33 & 86.70 & 71.21 \\
\bottomrule
\end{tabularx}
\end{table*}

\section{Additional Efficiency Results and Protocols}
\label{app:efficiency_protocols}

\subsection{Complete Measurements}
\label{app:efficiency_results}

Each backend runs one request per configuration after matched warmup and cache reset on one H20 with BF16 KV and elastic allocation. Greedy decoding generates 163,840 tokens from 8,192 input tokens with EOS stopping disabled and no live tool calls. Tables~\ref{tab:elastic_three_budgets} and~\ref{tab:elastic_vllm_three_budgets} report active KV, peak allocated GPU memory, and end-to-end latency. Ratios use each backend's own FullKV. SGLang uses FlashAttention3 for FullKV and Triton for compression. vLLM uses FlashAttention3 throughout.

\begingroup
\setlength{\intextsep}{4pt plus 1pt minus 1pt}
\renewcommand{\arraystretch}{.90}
\begin{table}[H]
\centering\small
\setlength{\tabcolsep}{2pt}
\caption{\textbf{SGLang efficiency.}}
\label{tab:elastic_three_budgets}
\begin{tabular}{@{}lrrrrrr@{}}
\toprule
Method & Budget & KV & Peak & Time & Mem. & Speed \\
 & tokens & GiB & GiB & s & ratio & ratio \\
\midrule
FullKV & -- & 23.625 & 41.43 & 2253.2 & 1.000 & 1.000\\
SnapKV & 256 & 0.053 & 18.56 & 993.3 & 2.232 & 2.268\\
SnapKV & 512 & 0.088 & 18.56 & 1013.2 & 2.232 & 2.224\\
SnapKV & 1024 & 0.158 & 18.56 & 1023.1 & 2.232 & 2.202\\
R-KV & 256 & 0.053 & 18.66 & 1056.7 & 2.220 & 2.132\\
R-KV & 512 & 0.088 & 18.66 & 1110.7 & 2.220 & 2.029\\
R-KV & 1024 & 0.158 & 18.66 & 1140.0 & 2.220 & 1.976\\
TriAttention & 256 & 0.053 & 18.93 & 1000.9 & 2.189 & 2.251\\
TriAttention & 512 & 0.088 & 18.93 & 984.2 & 2.189 & 2.289\\
TriAttention & 1024 & 0.158 & 18.93 & 1018.9 & 2.189 & 2.211\\
AgentKV & 256 & 0.053 & 18.58 & 1044.2 & 2.230 & 2.158\\
AgentKV & 512 & 0.088 & 18.58 & 1065.0 & 2.230 & 2.116\\
AgentKV & 1024 & 0.158 & 18.58 & 1069.9 & 2.230 & 2.106\\
\textbf{CommitKV} & 256 & 0.053 & 18.55 & 1060.3 & 2.233 & 2.125\\
\textbf{CommitKV} & 512 & 0.088 & 18.55 & 1072.9 & 2.233 & 2.100\\
\textbf{CommitKV} & 1024 & 0.158 & 18.55 & 1102.5 & 2.233 & 2.044\\
\bottomrule
\end{tabular}
\end{table}

\begin{table}[H]
\centering\small
\setlength{\tabcolsep}{2pt}
\caption{\textbf{vLLM efficiency.}}
\label{tab:elastic_vllm_three_budgets}
\begin{tabular}{@{}lrrrrrr@{}}
\toprule
Method & Budget & KV & Peak & Time & Mem. & Speed \\
 & tokens & GiB & GiB & s & ratio & ratio \\
\midrule
FullKV & -- & 23.625 & 41.20 & 2274.2 & 1.000 & 1.000\\
SnapKV & 256 & 0.053 & 18.82 & 1017.0 & 2.189 & 2.236\\
SnapKV & 512 & 0.088 & 18.83 & 986.0 & 2.189 & 2.307\\
SnapKV & 1024 & 0.158 & 18.83 & 1030.5 & 2.189 & 2.207\\
R-KV & 256 & 0.053 & 18.94 & 1061.5 & 2.175 & 2.142\\
R-KV & 512 & 0.088 & 18.94 & 1033.3 & 2.175 & 2.201\\
R-KV & 1024 & 0.158 & 18.95 & 1085.5 & 2.175 & 2.095\\
TriAttention & 256 & 0.053 & 18.76 & 1013.5 & 2.197 & 2.244\\
TriAttention & 512 & 0.088 & 18.76 & 1007.4 & 2.197 & 2.258\\
TriAttention & 1024 & 0.158 & 18.76 & 1005.7 & 2.197 & 2.261\\
AgentKV & 256 & 0.053 & 18.76 & 1070.0 & 2.196 & 2.126\\
AgentKV & 512 & 0.088 & 18.76 & 1053.2 & 2.196 & 2.159\\
AgentKV & 1024 & 0.158 & 18.76 & 1068.4 & 2.196 & 2.129\\
\textbf{CommitKV} & 256 & 0.053 & 18.80 & 1046.3 & 2.191 & 2.174\\
\textbf{CommitKV} & 512 & 0.088 & 18.80 & 1040.5 & 2.191 & 2.186\\
\textbf{CommitKV} & 1024 & 0.158 & 18.80 & 1063.1 & 2.191 & 2.139\\
\bottomrule
\end{tabular}
\end{table}
\endgroup

Earlier elastic SGLang runs at $B=512$ used 8,192 input tokens and generated either 131,072 or 163,840 tokens. CommitKV achieved $2.12\times$ and $2.45\times$ speedups, respectively, against the FullKV run at each length. The two lengths used separate frozen histories and each method pair shared its input. These runs used an earlier allocator configuration.

\section{Hyperparameter Sensitivity}
\label{sec:hyperparam_ablation}

\subsection{Fusion Weight on Disjoint Queries}
\label{app:frozen_fidelity}

In 32 probes from 16 trajectories, eight queries select tokens and eight disjoint queries measure output error against FullKV. All variants retain eight recent tokens and disable page-level reservations.

\begin{samepage}
\subsection{Task-Level Hyperparameter Ablations}
\label{app:task_hyperparameter_sensitivity}

\nopagebreak[4]
\begin{table}[H]
\centering\footnotesize
\setlength{\tabcolsep}{4pt}
\renewcommand{\arraystretch}{1.02}
\caption{\textbf{One-factor ablations on Qwen3-VL-8B-Instruct at $B=512$.} Scores are percentages. MMVP uses pair accuracy. The default unweighted mean is 61.97\%. Shading marks defaults and bold marks each sweep's column maxima.}
\label{tab:hyperparameter_sensitivity}
\begin{tabular}{@{}llrrrr@{}}
\toprule
Parameter & Value & MMSearch & V*Bench & MMVP & Avg. \\
\midrule
$\rho$ & 0 & 29.67 & 86.91 & 60.00 & 58.86 \\
$\rho$ & 0.125 & 31.67 & 88.48 & 59.33 & 59.83 \\
\rowcolor{kvsoftcyan}
$\rho$ & 0.25 (default) & \textbf{34.67} & \textbf{90.58} & \textbf{60.67} & \textbf{61.97} \\
$\rho$ & 0.5 & 29.67 & 90.05 & 60.00 & 59.91 \\
\midrule
$\lambda$ & 0 & 28.33 & 84.82 & 59.33 & 57.49 \\
$\lambda$ & 0.125 & 30.67 & 89.01 & 58.67 & 59.45 \\
\rowcolor{kvsoftcyan}
$\lambda$ & 0.25 (default) & \textbf{34.67} & \textbf{90.58} & \textbf{60.67} & \textbf{61.97} \\
$\lambda$ & 0.5 & 33.00 & 90.05 & 58.67 & 60.57 \\
\midrule
$\tau_{\mathrm{dead}}$ & 0.001 & 34.33 & \textbf{90.58} & 60.00 & 61.64 \\
$\tau_{\mathrm{dead}}$ & 0.005 & 32.00 & 88.48 & 60.00 & 60.16 \\
\rowcolor{kvsoftcyan}
$\tau_{\mathrm{dead}}$ & 0.01 (default) & \textbf{34.67} & \textbf{90.58} & 60.67 & \textbf{61.97} \\
$\tau_{\mathrm{dead}}$ & 0.02 & 30.67 & 88.48 & 60.00 & 59.72 \\
$\tau_{\mathrm{dead}}$ & 0.04 & 31.33 & 87.96 & \textbf{62.67} & 60.65 \\
\midrule
$\tau_{\mathrm{use}}$ & 0.02 & 30.67 & 87.43 & 60.00 & 59.37 \\
$\tau_{\mathrm{use}}$ & 0.05 & 31.67 & 88.48 & 59.33 & 59.83 \\
\rowcolor{kvsoftcyan}
$\tau_{\mathrm{use}}$ & 0.1 (default) & \textbf{34.67} & \textbf{90.58} & \textbf{60.67} & \textbf{61.97} \\
$\tau_{\mathrm{use}}$ & 0.15 & 34.00 & 89.01 & 60.00 & 61.00 \\
$\tau_{\mathrm{use}}$ & 0.2 & 32.33 & 89.53 & \textbf{60.67} & 60.84 \\
\midrule
$W$ & 4 & 31.33 & 88.48 & \textbf{61.33} & 60.38 \\
\rowcolor{kvsoftcyan}
$W$ & 8 (default) & \textbf{34.67} & \textbf{90.58} & 60.67 & \textbf{61.97} \\
$W$ & 16 & 31.67 & 90.05 & 58.00 & 59.91 \\
\midrule
$G$ & 8 & 32.00 & 89.53 & \textbf{61.33} & 60.95 \\
\rowcolor{kvsoftcyan}
$G$ & 16 (default) & \textbf{34.67} & \textbf{90.58} & 60.67 & \textbf{61.97} \\
$G$ & 24 & 34.00 & 90.05 & 60.67 & 61.57 \\
$G$ & 32 & 33.33 & 89.01 & \textbf{61.33} & 61.22 \\
\bottomrule
\end{tabular}
\end{table}
\end{samepage}

\section{More Analysis}
\label{app:observation_analyses}

\begin{samepage}
\subsection{Post-State-Matched Validation of Transition Evidence}
\label{sec:matched_transition_validation}

We match groups within each dataset by post-commit influence. Later reuse reaches 18.8\% for high--low pages and 10.2\% for low--low pages (Figure~\ref{fig:post_state_matched_controls}).

\nopagebreak[4]
\begin{figure}[H]
\centering
\includegraphics[width=\columnwidth]{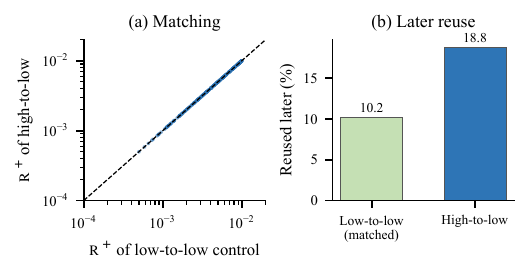}
\caption{\textbf{Post-state-matched control.} (a) Post-commit deletion effect of each high-to-low pair and its matched low-to-low control. (b) Fraction of each group reused in a later decoding window.}
\label{fig:post_state_matched_controls}
\end{figure}
\end{samepage}

\subsection{Extended Observation Analyses}
\label{sec:extended_observations}

\label{sec:direct_transition_validation}

We replay 60 Qwen3-VL-8B-Instruct FullKV trajectories with at least two calls, including ten answered correctly by FullKV and missed by SnapKV. Reuse means a later deletion effect above 0.05 at the last full-attention layer.

Under the SnapKV scoring rule at $B=512$, we mark a page as evicted when half its tokens are removed. Later reuse is 28.2\% (95\% interval: 24.3--33.1\%). On call and observation pages, low--low and high--low reuse is 3.2\% and 20.4\%, respectively (Figure~\ref{fig:observation}), using the original thresholds of 0.01 and 0.05.

\subsection{Additional Diagnostic Checks}
\label{app:mass_effect}
\label{app:new_transitions}

For Figure~\ref{fig:observation}d, the prediction target is whether a page is unused in all later recorded windows. We fit logistic regression to the log pre- and post-commit deletion effects and evaluate it with five-fold cross-validation grouped by trajectory. The controls rank pages by the post-commit effect or by attention alone.

The grouped score agrees with explicit deletion within $9.89\times10^{-7}$ across 192 checks.

\section{Qualitative Visual Examples}
\label{app:qualitative_examples}

\begin{figure*}[!tp]
\centering
\includegraphics[width=.96\textwidth]{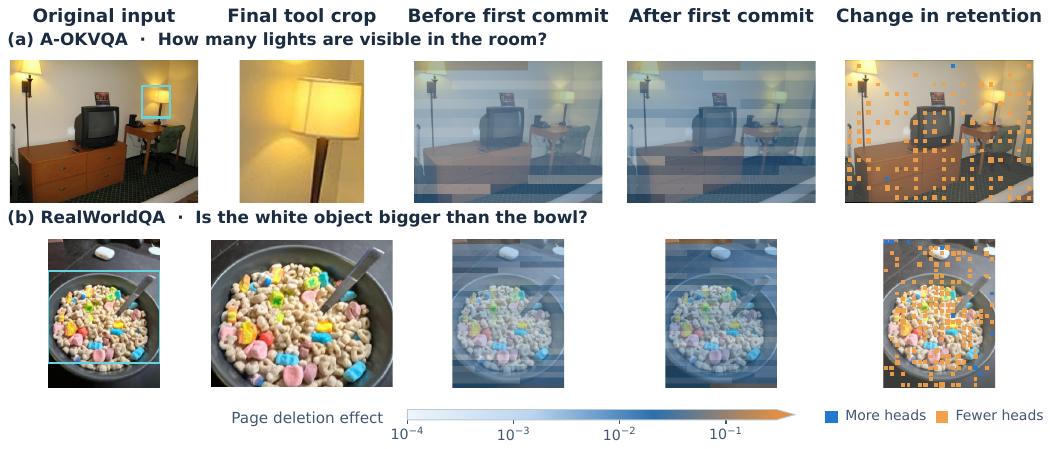}
\caption{\textbf{Deletion effects and token retention in two visual tasks.} The A-OKVQA example counts lights in a room. The RealWorldQA example compares a white object with a bowl of cereal. The box in each original image marks the final tool crop. Heatmaps show page deletion effects before and after the first commit on a shared log scale. Blue and orange mark tokens retained by more or fewer KV heads, respectively, when adding singleton deletion effects ($\lambda=0.25$, $B=512$).}
\label{fig:appendix_visual_cases}

\par\medskip
\includegraphics[width=.96\textwidth]{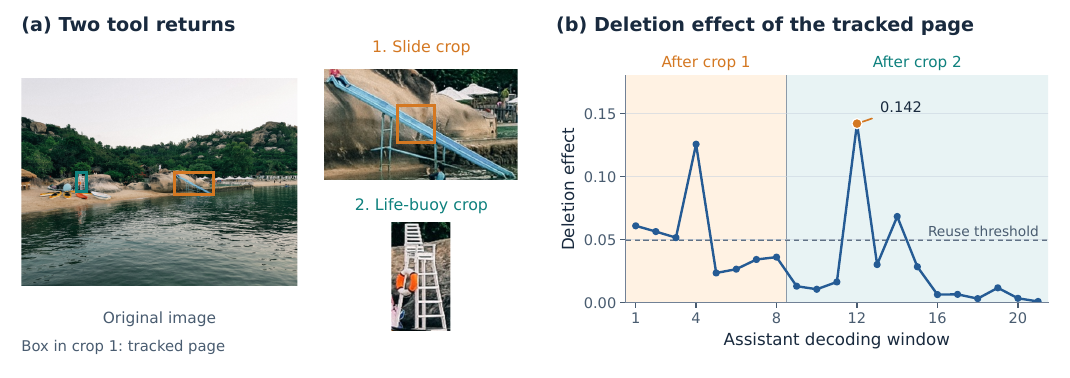}
\caption{\textbf{Delayed reuse across tool returns.} The box in the first crop marks the tracked visual page. The curve gives its deletion effect in windows of up to eight assistant queries at the last full-attention layer during FullKV replay. The dashed line marks the 0.05 reuse threshold.}
\label{fig:appendix_delayed_reuse}
\end{figure*}

Figure~\ref{fig:appendix_visual_cases} shows page influence and retention in two examples from A-OKVQA and RealWorldQA. Figure~\ref{fig:appendix_delayed_reuse} follows a V*Bench trajectory that asks where a slide is relative to a life buoy. The two tool crops cover disjoint regions. After the life-buoy crop returns, the slide page's deletion effect rises from 0.013 in the first decoding window to 0.142 in the fourth. The logged answer places the slide to the right of the life buoy.

\end{document}